\documentclass[11pt,a4paper]{article}
\usepackage{authblk}
\usepackage[hyperref]{acl2020}
\usepackage{times}
\usepackage{latexsym}

\usepackage{todonotes}
\usepackage{enumitem}
\usepackage{amssymb}
\usepackage{tabularx}
\usepackage{subcaption}
\usepackage{booktabs}
\usepackage{multirow}

\usepackage{microtype}

\aclfinalcopy 
\def\aclpaperid{1603} 

\newcommand\CORPUSNAME{OTTA}

\title{A Methodology for Creating Question Answering Corpora \\
Using Inverse Data Annotation}

\author[1]{Jan Deriu}
\author[1]{Katsiaryna Mlynchyk}
\author[1]{Philippe Schl{\"a}pfer}
\author[2]{Alvaro Rodrigo}
\author[1]{\\ Dirk von Gr{\"u}nigen} 
\author[1]{Nicolas Kaiser}
\author[1]{Kurt Stockinger}
\author[3]{Eneko Agirre}
\author[1]{Mark Cieliebak}

\affil[1]{Zurich University of Applied Sciences (ZHAW), Winterthur, Switzerland}
\affil[ ]{\textit {\{deri, mlyn, scrp, vogr, stog, ciel\}@zhaw.ch}}
\affil[2]{National Distance Education University (UNED), Madrid, Spain}
\affil[ ]{\textit {alvarory@lsi.uned.es}}
\affil[3]{University of the Basque Country (UPV/EHU), Donostia, Spain}
\affil[ ]{\textit {e.agirre@ehu.eus}}

\date{}

\begin{document}
\maketitle
\begin{abstract}
In this paper, we introduce a novel methodology to efficiently construct a corpus for question answering over structured data. For this, we introduce an intermediate representation that is based on the logical query plan in a database called Operation Trees (OT). This representation allows us to invert the annotation process without losing flexibility in the types of queries that we generate. Furthermore, it allows for fine-grained alignment of query tokens to OT operations. \newline
In our method, we randomly generate OTs from a context-free grammar. Afterwards, annotators have to write the appropriate natural language question that is represented by the OT. Finally, the annotators assign the tokens to the OT operations. We apply the method to create a new corpus  \textit{\CORPUSNAME} (Operation Trees and Token Assignment), a large semantic parsing corpus for evaluating natural language interfaces to databases. We compare \CORPUSNAME~ to Spider and LC-QuaD 2.0 and show that our methodology more than triples the annotation speed while maintaining the complexity of the queries. Finally, we train a state-of-the-art semantic parsing model on our data and show that our corpus is a challenging dataset and that the token alignment can be leveraged to increase the performance significantly.
\end{abstract}

\section{Introduction}
Question Answering (QA) over structured data, also called Natural Language Interfaces to Databases (NLI2DB) or Text-to-SQL, is a key task in natural language processing and the semantic web. It is usually approached by mapping a natural language question (NL question) into executable queries in formal representations such as logical forms, SPARQL or SQL. 

The state-of-the-art in this problem uses machine learning techniques to learn the mapping. Unfortunately, the construction of labeled corpora to train and evaluate NLI2DB systems is time- and cost-intensive, which is slowing down progress in this area. In particular, it usually requires recruiting SQL or SPARQL experts to write queries for natural language questions.   For instance, in Spider \cite{yu-etal-2018-spider}, the authors recruited students to write SQL queries.  They worked 500 person-hours to generate 5,600 queries, which corresponds to more than 5 minutes per question. 

As a more cost-effective alternative to writing formal queries manually, some authors propose to use templates to generate them automatically. For instance, LC-QUAD 2.0 \cite{dubey2019lc} used 22 templates based on the structure of the target knowledge graph. Constructing templates is also time-consuming, and the expressiveness of the automatically produced queries is limited. 


Apart from the high cost of generating queries, the natural
language questions in current datasets do not necessarily cover the whole range of data present in the database. In Spider, the coverage is limited by the creativity of the students, and in LC-QUAD 2.0 by the templates.

In this paper, we propose a new procedure to increase the speed of the annotation process. For this, we first introduce an intermediate representation of the structured queries, which we call \emph{Operation Trees} (OTs, see Figure \ref{fig:example1}). Our OTs follow a context-free grammar and are based on logical query plans that can easily be mapped to SPARQL or SQL, making our system more versatile. In addition, it has been shown that working on abstract tree representations instead of sequences yields better results \cite{guo-etal-2019-towards}. Recent work by \cite{cheng2019learning} shows the successful use of tree-like abstractions as an intermediate representation to parse text into semantic representations, reinforcing our choice of operation trees as the main representation language. 

Our annotation process works as follows. First, we use the context-free grammar to sample random OTs for a given database. We then let annotators in a first round write the corresponding NL questions for the sampled OTs. In a second, optional, round the annotators perform an assignment of tokens from the NL question to the operations in the OT. This additional annotation enriches the information in the dataset, and, as we will show below, allows for performance gains, especially in low data regimes.

Our approach to producing datasets has the following advantages with respect to the methodology used in previous work: 1) It reduces the time needed for an annotation (less than 2 minutes, compared to more than 5 in Spider). 2) It allows us to cover the whole range of data present in the database structure and not to focus on the most prominent examples. 3) Our annotation procedure provides alignments between operations in the formal language and words in the question, which are an additional source of supervision when training. 

We applied our approach\footnote{The annotation tool can be found here: \url{https://github.zhaw.ch/semql/annotation_tool}}~to five datasets, yielding a large corpus called \CORPUSNAME\footnote{The corpus can be found here: \url{https://github.zhaw.ch/semql/semql-data/tree/master/annotated_tree_files/single_files}}~which consists of 3,792 complex NL questions plus their corresponding OTs as well as the token assignment for one of our domains. Besides, we have adapted a state-of-the-art system  \cite{yin-neubig-2017-syntactic} to work on to operation trees, and included a mechanism to profit from token alignment annotations when training. The system yields better results with up to 7 point increase when trained on aligned OTs. 


\section{Related Work}


In this section, we first review the related work in the area of Natural Language Interfaces to Databases (NLI2DB). Afterwards, we focus on the data resources that are currently available to evaluate these systems.

\paragraph{Natural Language Interfaces to Databases.}
There is a vast amount of literature on NLI2DB. A recent survey on methods and technologies is provided by \cite{Affolter2019}. Early systems use a {\it keyword-based} approach with inverted indexes to query the databases \cite{simitsis2008precis, blunschi2012soda, bast2015more}. {\it Pattern-based} approaches are able to handle more complex NL questions \cite{damljanovic2010natural, zheng2017natural}. {\it Parsing-based} approaches use a natural language parser to analyze and reason about the grammatical structure of a query \cite{li2014constructing, saha2016athena}. {\it Grammar-based} approaches only allow the user to formulate queries according to certain pre-defined rules, thus focus primarily on increasing the precision of answers \cite{song2015tr, ferre2017sparklis}. More recent systems use a {\it neural machine translation} approach similar to translating natural languages, say, from French to English \cite{iyer2017learning, basik2018dbpal, cheng2019learning, liu2019leveraging, guo-etal-2019-towards, cheng2019learning}. 

\paragraph{Data Resources.}
We will now review the major data resources that have recently been used for evaluating NLI2DB systems. These resources are mainly created following two approaches: (1) Both NL and structured queries are manually created, and (2) structured queries are automatically generated, and then humans create the corresponding NL questions.

Regarding fully manually created resources, \cite{yu-etal-2018-spider} provided Spider, a dataset with 5,600 SQL queries, over 200 databases and 10,181 NL questions annotated by 11 students, where some questions were manually paraphrased to increase the variability. \cite{acl18sql} released Advising, with 4.5k questions about university course advising and SQL queries. \cite{Dahl:1994:ESA:1075812.1075823} created ATIS, a dataset with 5k user questions about flight-booking manually annotated with SQL queries and modified by \cite{iyer-etal-2017-learning} to reduce nesting. \cite{Zelle:1996:LPD:1864519.1864543} created GeoQuery, with 877 questions about US geography annotated with Prolog and converted to SQL by \cite{Popescu:2003:TTN:604045.604070} and \cite{Giordani:2012:AGR:3069470.3069477}. 
There are also smaller datasets about restaurants with 378 questions \cite{tang-mooney-2000-automated}, the Yelp website with 196 questions and IMDB with 131 questions \cite{Yaghmazadeh:2017:SQS:3152284.3133887}.




Resources using an automatic step usually rely on generating structured queries using templates created by experts. \cite{zhongSeq2SQL2017} created WikiSQL, a collection of 80k pairs of SQL queries and NL questions made using Wikipedia. 
However, SQL queries are relatively simple because each of the databases consists of only a single table without foreign keys. Hence, the queries do not contain joins. \cite{dubey2019lc} developed LC-QuAD 2.0, with 30,000 complex NL questions and SPARQL queries over DBpedia and Wikidata. They used templates to generate SPARQL queries for seed entities and relations, which are lexicalized automatically using other templates. NL questions of both datasets were created by crowdsourcing workers.

All the resources mentioned above required a large amount of effort. In each case, the annotators need an in-depth knowledge of SQL or a similarly structured language. Our approach simplifies the process of generating question-answering corpora while ensuring a large coverage of the underlying database without forfeiting any complexity in the queries.

On the other hand, \citep{wang-etal-2015-building} developed a method similar to ours. They begin with a lexicon linking natural utterances with predicates in the database. Then, they use domain-specific grammar to create several canonical phrases associated with queries. Finally, crowdsourcing workers rewrite the canonical phrases and create natural utterances used for training a semantic parser. Similar to our approach, they combine an automatic method with crowdsourcing workers. However, they have to create the lexicon and the grammar for each database, while our method can be applied to any database without creating new resources.



\section{Operation Trees}

\begin{figure}[h!]
	\begin{center}
        \begin{tabular}{@{}c@{}}
		\includegraphics[width=0.32\textwidth]{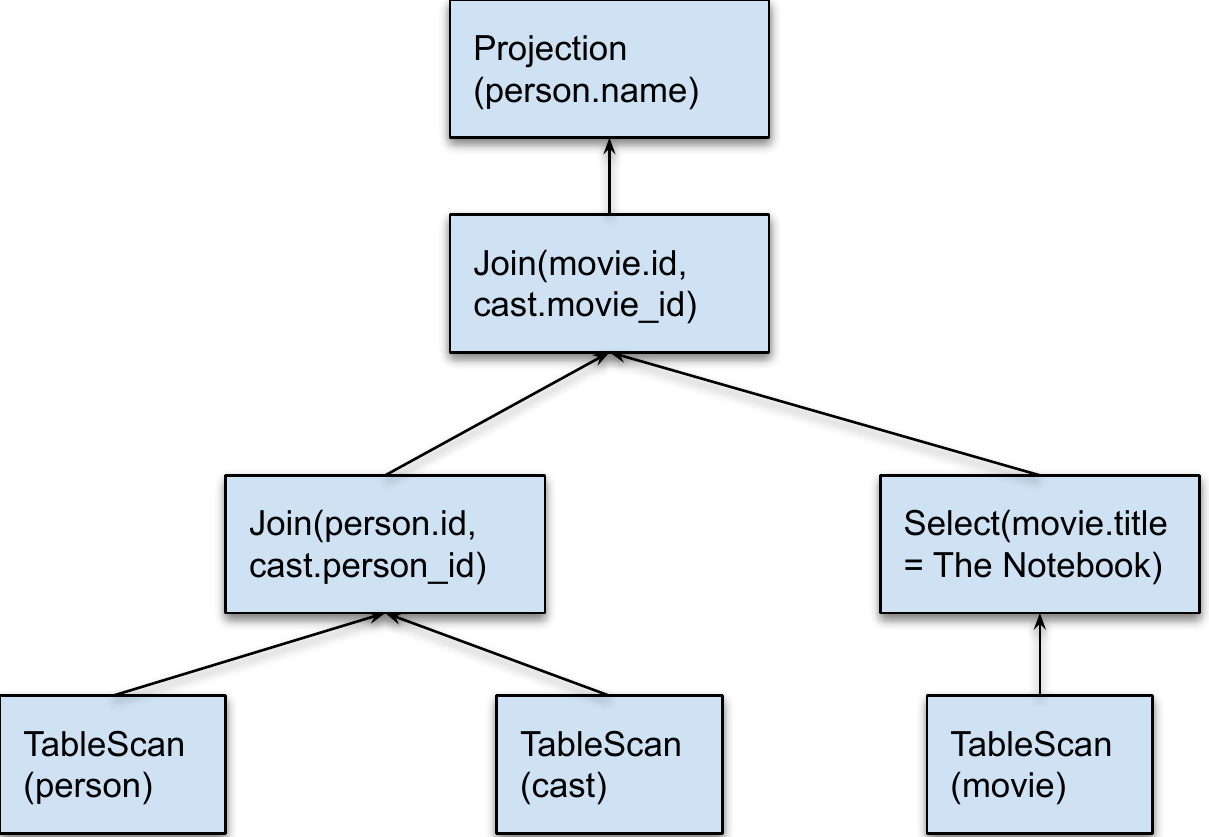} \\ 
		(a) \vspace{2mm}\\
		\includegraphics[width=0.32\textwidth]{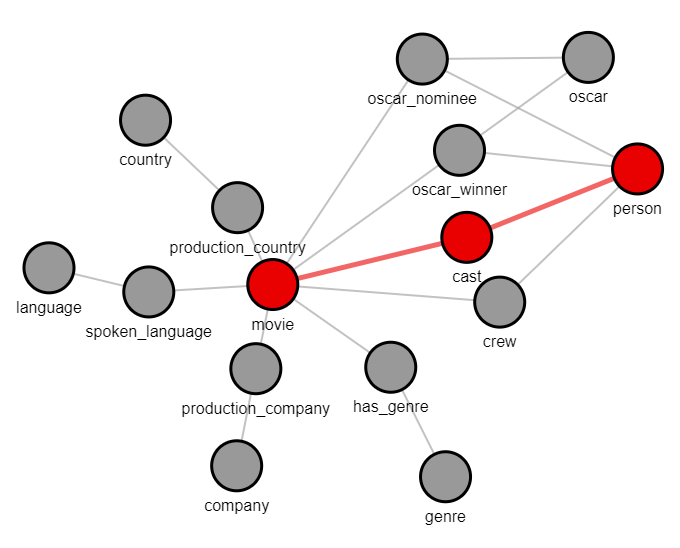} \\ 
		(b) \\
       \end{tabular}
	\end{center}\vspace{-3mm}
	\captionof{figure}{(a) Example of an Operation Tree (OT) for the query "Who starred in 'The Notebook'?" (b) The corresponding database schema.}
\label{fig:example1}
\end{figure}

In our setting, the goal is to generate an Operation Tree (OT) that finds the correct answer for a given question in natural language. An OT is a binary tree that is closely related to a logical query plan in SQL database engines. An OT is composed of a sequence of operations that can be mapped to a database query language such as SQL or SPARQL to retrieve the proper result. 

\noindent {\bf Example.}
Assume that we have a database about movies that we want to query in natural language. In Figure \ref{fig:example1}, an example of an OT is depicted for the question "Who starred in 'The Notebook'?". In order to answer this question, the tables \emph{person} and \emph{movie} are selected, then the table \emph{movie} is filtered by movie title \emph{The Notebook}. In the next step, the tables are joined via the bridge-table \emph{cast}. Finally, the \emph{person.name} column is extracted.

We enhance these OTs by associating a reasonable subset of tokens from the NL question to each operation in the tree. For instance, the token "starred" could be associated to the \emph{Join} operation, as this operation implies that an actor starred in a movie, whereas the tokens "How many" could be associated to the \emph{Count} operation. This mapping between tokens and operations will help later on to train machine learning algorithms to generate OTs automatically from natural language questions with better quality.  

\begin{figure}[h!]
	\begin{center}
		\includegraphics[width=0.45\textwidth]{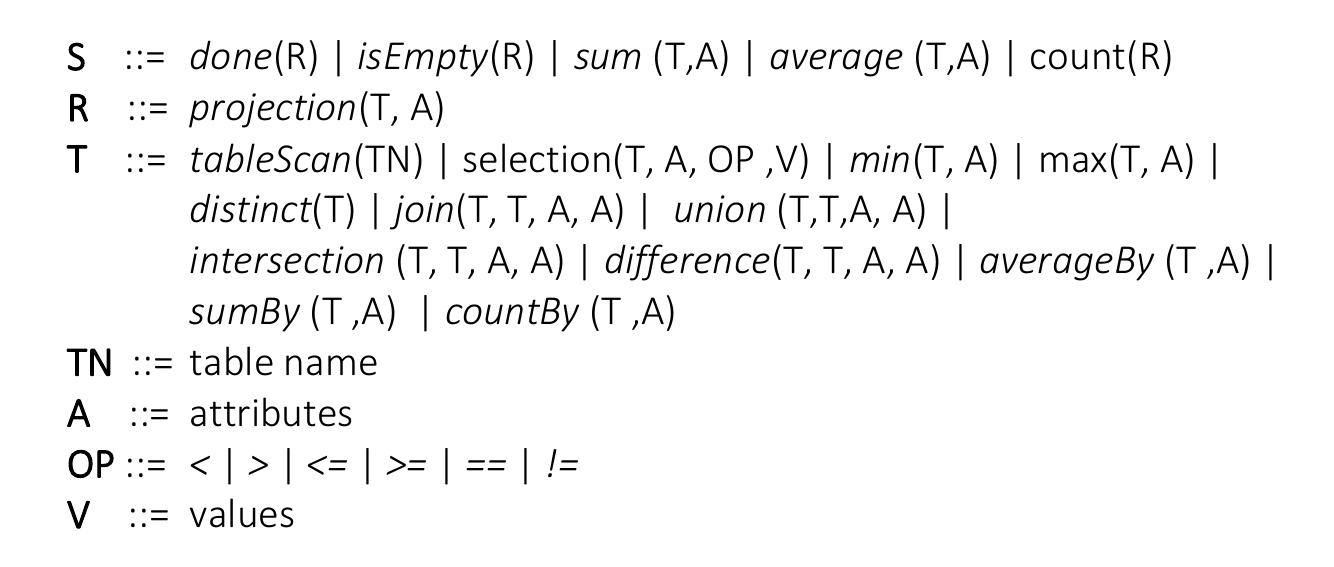}
	\end{center}
	\caption{The set of production rules for the context-free grammar of the operation trees, where \emph{table name} denotes the set of all entity types in the database, \emph{attributes} denotes the set of all attributes of entity types, and \emph{values} denotes the set of all entries in the database. The non-terminal symbols \textbf{S}, \textbf{T},and \textbf{R} denote the start-symbol, intermediate tables, and result tables respectively. }
\label{fig:grammar}
\end{figure}

\noindent {\bf Definition.} More formally, the OTs follow a predefined context-free grammar. In the current state, the set of operations includes major operations from the relational algebra with specific extensions. The full grammar is shown in Figure \ref{fig:grammar}. 

The OTs can be used to represent queries for any entity-relationship data paradigm. For instance, in SQL databases the entity types are the tables, the attributes are the columns, and the relationships are represented as tables as well. Similar mapping is possible for other paradigms. 

\noindent {\bf Properties.}
The OTs have several features:
\begin{itemize}[noitemsep,leftmargin=*,label={}]
    \item {\it Question Types}: There are different types of questions that can be asked. For instance, 1) yes/no questions (\emph{IsEmpty}), 2) questions about a list of items (\emph{Projection} followed by \emph{Done}), 3) questions about the cardinality of a result set (\emph{Count}), and 4) questions about an aggregation (\emph{Sum, Avg,} etc.). 
    \item {\it Result Types}: The type of results is defined by the entity types in the result set. For instance, a question can ask about the list of directors that satisfy certain constraints (e.g., all directors that were born in France). In this case, the result type would be the \emph{person} type. 
    \item {\it Constraints}: The constraints represent the filters that are applied onto the attributes of the entities. For instance, "All directors born in France" sets a constraint on the \emph{birth\_place} attribute.   
    \item {\it Entity Types}: They define which entity types are involved in the query. The selected entity types are combined, usually via a \emph{Join} operation. For instance, in Figure \ref{fig:example1} the entity types are \emph{movie} and \emph{person}, which are combined with the table \emph{cast}.
    \item {\it Aggregation Types:} They define reduction operations, which are applied to the data. This includes \emph{Min/Max operations} on an attribute, \emph{Set operations} on two sets of relations, and \emph{Group By operations}. 
\end{itemize}

\noindent {\bf Complexity.} In order to categorize the OTs, we define a complexity score similar to \cite{yu-etal-2018-spider}, which is based on the number of components in the tree. The more \emph{Joins} and \emph{Group By} operations, \emph{Aggregations} or \emph{Filters} are in the query, the higher the score. Like  \cite{yu-etal-2018-spider}, we define four categories: \emph{Easy}, \emph{Medium}, \emph{Hard}, and \emph{Extra Hard}. 


\section{Corpus Construction}
The evident way to construct a corpus with NL questions and their corresponding OT queries would consist of two main parts: first, collect a set of NL questions, and then create the corresponding OT queries to these questions. However, this approach is very time-consuming and has a major issue. In essence, questions tend to be very narrow in scope, i.e., they do not necessarily cover the whole range of entity types, attributes and relationships that are present in the database. Moreover, writing the corresponding OT queries for the NL questions requires sufficient SQL skills as well as a mechanism to verify that the OT statements actually correspond to the question. 

Thus, we decided to \emph{invert the process}. That is, we first randomly sample an OT using the above-defined context-free grammar, and then annotators write a corresponding question in natural language. In the last step, annotators manually map tokens of the question to the operations. 
There are several advantages to this procedure: 1) It allows for controlling the characteristics of the OTs, i.e., we can control the question type, the response type, the constraints, and the entity type. 2) It allows them to create more complex questions that better cover the variety of the underlying data. 3) The annotation process is less time consuming, as the annotators do not have to build the trees or write queries. Rather they can focus on writing the question and assigning tokens.
We now describe the process of automatic sampling and manual annotation in more detail. 
\subsection{Tree Sampling}
The tree sampling procedure is composed of the following steps:
\begin{itemize}[noitemsep,leftmargin=*,label={}]
    \item {\it Question Type}: This can be sampled at random or be manually set if a certain type is desired.
    \item {\it Result Type}: First, an entity type is randomly sampled. Then a specific set of attributes is sampled from the chosen entity type. Alternatively, the result type can be manually set.
    \item {\it Entity Types}: The entity types are sampled based on the graph structure of the entities and relationships in the database schema. For this, we sample from all the possible join-paths, which contain the table of the result type. This is also controllable, as we can specify the length of the paths we want to consider. 
    \item {\it Constraints}: In the constraints, the filter arguments are sampled. First, the entity types are randomly selected on which the constraints are to be applied. Then we sample an operation and a value at random for each entity type and each attribute. We can limit the number of overall constraints and the number of maximum constraints for each entity type. 
    \item {\it Group By}: The \emph{Group By} operations (\emph{AvgBy, SumBy, CountBy}) are chosen at random. For a \emph{Group By} operation, two attributes need to be selected: a group-attribute, which defines on which attribute to group, and an aggregation-attribute, which defines on which column to apply the aggregation. For instance, we could group by genre and aggregate over the movie budget. 
    \item {\it Tree structure}: The tree structure is sampled as follows. First, the \emph{Join} operations are applied on the sampled entity types. Second, the set operations (\emph{Union, Intersect, Diff}) are inserted. Third, the \emph{Selection} operations are inserted. Next, the aggregation operations are inserted, i.e., \emph{Group By, Min, Max} operations. Finally, the operations for the question type are sampled. For instance, if the question type is a list of entities, then we use the \emph{Projection} operation, but if it is a cardinality question, we use the \emph{Count} operation.  
\end{itemize}
This procedure may create trees that make no sense semantically. We handle those trees during the annotation phase, which we describe below. Furthermore, we make sure that the trees are executable. For this, we translate the trees into SQL and run them on the database. We also omit trees that return an empty result, as they can lead to confusions during the evaluation, as two different queries that both return an empty result would be counted as being equal. 
\subsection{Annotation}
The annotation process, i.e., writing natural language questions and assigning query tokens to operations in the OT, is performed in two phases. For each phase, we developed a graphical user interface to facilitate the annotation process (for more details, see Appendix \ref{sec:appendix_annoation}). \\
\noindent {\bf Phase 1.} In the first phase, the annotator is presented with an OT, which is automatically sampled as described in the previous section. The task of the annotator is to formulate an appropriate NL question for the sampled OT. In some cases, the sampled tree has contradicting or nonsensical constraints (e.g., compute the average year). For these cases, the annotators can either skip or adapt the OT by changing the constraints. \\
\noindent {\bf Phase 2.} In the second phase, the annotators perform the token assignment as well as quality control. The annotators are presented with an OT and the NL question, which was written by a different annotator in phase 1. First, they check and correct the NL question, then they assign the tokens to the operations. In order to achieve consistent annotation results, we set up a guideline on how the tokens are to be assigned (more information in the Appendix).


\section{Corpus \CORPUSNAME}

We applied our corpus construction procedure to a set of five databases and produced a new corpus with NL questions and corresponding OTs, called \CORPUSNAME. In order to compare our results with previous work, we used four databases from the \emph{Spider} corpus (CHINOOK, COLLEGE, DRIVING SCHOOL, and FORMULA I), which we extended with a dump from IMDB\footnote{https://www.imdb.com/} that we refer to as MOVIEDATA. 
For the annotations, we employed 22 engineers with basic knowledge in SQL-databases. 

\subsection{Corpus Statistics}
Table \ref{table:data-overview} summarizes the dataset. The number of tables per database ranges from 6 to 18, and the number of attributes ranges from 45 to 93 columns per database. 
For CHINOOK and MOVIEDATA, our corpus has more than  1000 annotated OTs, while it has around 500 annotated OTs for the other three databases. For MOVIEDATA, we also performed the token annotation procedure.
For each database, we computed the average \emph{complexity} score. Except for MOVIEDATA, which is \emph{Hard}, all other databases have a \emph{Medium} average query complexity. 
The average time per question annotation ranges from 77 to 104 seconds (average 97.7 seconds). The token assignment and question correction, on the other hand, took on average 101 seconds per OT. 

\begin{table*}[h!]
\vspace{-1mm}
\begin{center}
\begin{small}
\resizebox{.95\textwidth}{!}{
\begin{tabular}{l||ccccc} 
\hline
& \textsc{MOVIEDATA} & \textsc{CHINOOK} & \textsc{COLLEGE} & \textsc{DRIVING SCHOOL} & \textsc{FORMULA1}\\
\hline 
\textsc{\#Tables}  & 18        & 11      & 11      & 6              & 13       \\
\textsc{\#Attributes} & 64        & 63      & 45      & 39             & 93       \\
\textsc{\#Queries}    & 1148      & 1067    & 462     & 547            & 568      \\
\textsc{Time per Annotation (sec)}     & 104      & 104    & 77     & 78            & 104 \\
\textsc{Avg. Complexity}     & Hard      & Medium    & Medium     & Medium            & Medium \\
\hline
\end{tabular}
}
\end{small}
\\~\\
  \vspace{1mm}
\end{center}
\caption{Statistics of the new corpus \CORPUSNAME}
\label{table:data-overview}
\end{table*}

\begin{table*}[h!]
\vspace{-1mm}
\begin{center}
\begin{small}
\resizebox{.95\textwidth}{!}{
\begin{tabular}{l||ccccccccc} 
\hline
& \textsc{\#Questions} & \textsc{\#Queries} & \textsc{\#DB} & \textsc{\#Table/DB} & \textsc{Table Cov.} & \textsc{Attr Cov.} & \textsc{MSTTR} & \textsc{Avg. \#Tokens} & \textsc{Ann. Time}\\
\hline 
\textsc{Spider} &  10,181        & 5,693      & 200      & 5.1 & 0.917 (0.87) & 0.621 (0.496) & 0.519 & 12.67 & 360 sec.     \\
\textsc{LC-QuaD 2.0}  &  30,000        & 30,000     & 1      & 157,068            & 0.019 & 0.187 & 0.761 & 10.6  & -      \\
\textsc{\CORPUSNAME~(ours)}   &  3,792       & 3,792      & 5      & 11.8 & 0.949 & 0.544 & 0.67 & 13.53 & 98 sec.    \\
\hline
\end{tabular}
}
\end{small}
\\~\\
  \vspace{1mm}
\end{center}
\caption{Comparison of our corpus \CORPUSNAME~to the Spider and LC-QuaD 2.0 corpora.  Note that the number of databases in LC-QuaD 2.0 is only 1, since it is an open-domain knowledge base, and the number of tables corresponds to the number of different classes. Numbers in parentheses only consider databases with more than 5 tables.}
\label{table:query-comparison}
\end{table*}

\begin{table*}[h!]
\vspace{-1mm}
\begin{center}
\begin{small}
\resizebox{.95\textwidth}{!}{
\begin{tabular}{l||cccccccc} 
\hline
& \textsc{\#Avg. Join} & \textsc{\#Group By} & \textsc{\#Order By} & \textsc{\#Nested} & \textsc{\#Having} & \textsc{\#Set Op} & \textsc{\#Aggregations} & \textsc{\#Boolean} \\
\hline 
\textsc{Spider} &  0.537 & 0.262 & 0.234  & 0.148  & 0.068 & 0.076 & 0.519 & -      \\
\textsc{LC-QuaD 2.0}  &  2.05 hops   &0 & 0.041 & 0  & 0 & 0 & 0.048  & 0.089    \\
\textsc{\CORPUSNAME~(ours)}   &  1.19  &  0.133 & 0      & 0      & 0.117     & 0.02 & 0.4  &  0.161   \\
\hline
\end{tabular}
}
\end{small}
\\~\\
  \vspace{1mm}
\end{center}
\caption{Comparison of the query complexity based on the ratio of components per query. For the aggregations in LC-QuaD 2.0, we report  the number of queries that use a \emph{Count} operation.}
\label{table:complexity-comparison}
\end{table*}

\subsection{Corpus Comparison}
In order to examine our corpus, we compare its characteristics to the \emph{Spider} corpus and to the \emph{LC-QuAD 2.0} corpus. We compare the coverage of the queried data, the complexity of the natural language questions and the complexity of the corresponding SPARQL/SQL queries.  

\noindent {\bf Coverage.}  Table \ref{table:query-comparison} shows the major characteristics of the three corpora. We compare the coverage of the databases in terms of the ratio of tables and attributes which appear in the queries. 

The average \emph{attribute coverage} of Spider over all databases equals $62.1\%$. However, more than half of the databases in Spider contain 5 tables or less. Thus, we also report the coverage of attributes only considering the databases which have more than 5 tables, where Spider only covers $49.6\%$ of attributes. Corpus \CORPUSNAME, in contrast, covers $54.4\%$ of all attributes. Furthermore, the divide becomes more apparent when we consider databases with larger amounts of tables. For instance, for the FORMULA-1 database, our corpus covers $44.2\%$ of all attributes, in contrast to Spider, where only $22.1\%$ of attributes are covered. LC-QuaD 2.0 covers 1,310 out of 7,005 properties\footnote{ For the number of classes and properties in Wikidata, we consulted: \url{https://tools.wmflabs.org/sqid}} (i.e. attributes in SQL), which corresponds to $18.7\%$. This is an extensive coverage, considering the high amount of properties.

The \emph{table coverage} shows a similar picture: our approach covers $94.9\%$ of all tables in the databases, whereas Spider covers $91.7\%$. This number drops down to $87\%$ when considering only databases with more than 5 tables. Again, this effect is most pronounced for the FORMULA-1 database, where we cover $92\%$ of the tables, whereas Spider only covers $69.2\%$. This shows that our method better scales to larger databases, which is relevant for real-world applications, where databases with a vast number of tables exist. LC-QuaD 2.0 covers around $1.9\%$ of approx. 160k classes, which makes comparison hard, as it is impossible to cover this vast amount of classes with 30k queries.



\noindent {\bf Query Complexity.} In order to compare the complexity of the queries, we examine the number of occurrences of different components in the queries (see Table \ref{table:complexity-comparison}). 

We first observe that our corpus \CORPUSNAME~does not contain any queries with \emph{Order By} operators or nested queries - however, they could be easily added to the grammar to fill this gap. Furthermore, Spider contains more aggregation operations (in particular \emph{Min, Max, Count, Average}, and \emph{Sum}). Again,  this could be easily adapted in our corpus by sampling more trees that contain these aggregations. On the other hand, our corpus stands out in the number of \emph{joins} per query: on average \CORPUSNAME~has 1.19 join operations per query in contrast to Spider, which has 0.537 joins per query. In fact, about $40\%$ of the queries in Spider contain joins, whereas \CORPUSNAME~is composed of $54\%$ of queries, which contain at least one join operation. Furthermore, around $37\%$ of our queries contain two joins in contrast to $9\%$ in Spider. On the other hand, \emph{LC-QuaD 2.0} contains an average of 2 hops (equivalent to two joins in relational databases) per query, which lies in the nature of graph database queries that are optimized for handling queries that range over multiple triple patterns. However, \emph{LC-QuaD 2.0} lacks complexity when considering more complex components (e.g., Group By, Set-Operation, etc.). In addition to the operations in relational algebra, the OTs also support \emph{Boolean} questions (i.e., yes/no questions), which make $16.1\%$ of our corpus compared to $8.9\%$ in LC-QuaD 2.0.

\noindent {\bf Question Complexity.}  The lexical complexity of the NL questions is measured in terms of \emph{mean-segmental token-type-ratio (MSTTR)} \cite{doi:10.1080/09296171003643098}, which computes the number of different token types in relation to all tokens in a corpus. The MSTTR is computed over text segments of equal length, in order to avoid biases due to different lengths within the corpora. 
First, note that the average length of the questions in all three corpora is approximately the same, between 10.6-13.6 tokens on average. Table \ref{table:query-comparison} shows that our corpus contains a much higher lexical complexity of the questions than  Spider (0.67 instead of 0.52). Thus, our approach seems to avoid trivial or monotonous questions, which also matches with our impression from manual inspection. On the other hand, the lexical complexity is higher in \emph{LC-QuaD 2.0}, which is due to the open domain nature of the dataset. 

\begin{table*}[h!]
\centering
\footnotesize
\resizebox{0.9\textwidth}{!}{\begin{minipage}{1\textwidth}
\begin{tabular}{|l|l|l|}
\hline
\textbf{Hardness} & \textbf{Spider} & \textbf{OTTA} \\ \hline
\multirow{1}{*}{\begin{minipage}{0.01\textwidth}easy\end{minipage}}
& \multirow{2}{*}{\begin{minipage}{0.45\textwidth}Find the number of albums.\end{minipage}}
& \multirow{2}{*}{\begin{minipage}{0.45\textwidth}Where were the invoices with the total sum of 1.99 or smaller issued? \end{minipage}}
\\ & & \\ 
\cline{2-3}
& \multirow{2}{*}{\begin{minipage}{0.45\textwidth}What is the average unit price of all the tracks?\end{minipage}}
& \multirow{2}{*}{\begin{minipage}{0.45\textwidth}What are the unit prices of tracks composed by Alfred Ellis/James Brown?\end{minipage}}
\\ & & \\ 
\cline{2-3}
& \multirow{2}{*}{\begin{minipage}{0.45\textwidth}Find all the customer information in state NY.\end{minipage}}
& \multirow{2}{*}{\begin{minipage}{0.45\textwidth}To which country belongs the 89503 postal code?\end{minipage}}
\\ & & \\ 
\cline{2-3}
\hline

\multirow{1}{*}{\begin{minipage}{0.01\textwidth}medium\end{minipage}}
& \multirow{2}{*}{\begin{minipage}{0.45\textwidth}Count the number of tracks that are part of the rock genre.\end{minipage}}
& \multirow{2}{*}{\begin{minipage}{0.45\textwidth}What is the average length of the tracks in the Grunge playlist?\end{minipage}}
\\ & & \\ 
\cline{2-3}
& \multirow{2}{*}{\begin{minipage}{0.45\textwidth}Please show the employee first names and ids of employees who serve at least 10 customers.\end{minipage}}
& \multirow{2}{*}{\begin{minipage}{0.45\textwidth}When did we sell tracks larger than 8675345 bytes?\end{minipage}}
\\ & & \\ 
\cline{2-3}
& \multirow{2}{*}{\begin{minipage}{0.45\textwidth}Find the name of the artist who made the album "Balls to the Wall".\end{minipage}}
& \multirow{2}{*}{\begin{minipage}{0.45\textwidth}To which postal codes did we sell a track named Headspace?\end{minipage}}
\\ & & \\ 
\cline{2-3}
\hline

\multirow{1}{*}{\begin{minipage}{0.01\textwidth}hard\end{minipage}}
& \multirow{2}{*}{\begin{minipage}{0.45\textwidth}What is the average duration in milliseconds of tracks that belong to Latin or Pop genre?\end{minipage}}
& \multirow{2}{*}{\begin{minipage}{0.45\textwidth}How many different playlists with a track that is bigger than 7045314 bytes do exist?\end{minipage}}
\\ & & \\ 
\cline{2-3}
& \multirow{2}{*}{\begin{minipage}{0.45\textwidth}What are the names of artists who have not released any albums?\end{minipage}}
& \multirow{2}{*}{\begin{minipage}{0.45\textwidth}What is the album title having the track with the lowest length in milliseconds in the genre name  Sci Fi \& Fantasy?\end{minipage}}
\\ & & \\ 
\cline{2-3}
& \multirow{2}{*}{\begin{minipage}{0.45\textwidth}What are the last names of customers without invoice totals exceeding 20?\end{minipage}}
& \multirow{2}{*}{\begin{minipage}{0.45\textwidth}What are the genres from artists not named Scholars Baroque Ensemble?\end{minipage}}
\\ & & \\ 
\cline{2-3}
\hline

\multirow{1}{*}{\begin{minipage}{0.01\textwidth}extra\end{minipage}}
& \multirow{2}{*}{\begin{minipage}{0.45\textwidth}What is the name of the media type that is least common across all tracks?\end{minipage}}
& \multirow{2}{*}{\begin{minipage}{0.45\textwidth}Whats the total unit price sold to customers with the email hholy@gmail.com and Argentina as billing country?\end{minipage}}
\\ & & \\ 
\cline{2-3}
& \multirow{2}{*}{\begin{minipage}{0.45\textwidth}Count the number of artists who have not released an album.\end{minipage}}
& \multirow{2}{*}{\begin{minipage}{0.45\textwidth}How many different genres do the tracks have, which were bought by customers who live in France?\end{minipage}}
\\ & & \\ 
\cline{2-3}
& \multirow{2}{*}{\begin{minipage}{0.45\textwidth}What are the album titles for albums containing both Reggae and Rock genre tracks?\end{minipage}}
& \multirow{2}{*}{\begin{minipage}{0.45\textwidth}Which customers made at least 35 purchases, excluding titles from the Chico Science \& Nacao Zumbi album?\end{minipage}}
\\ & & \\ 
\cline{2-3}
\hline

\end{tabular}

\caption{\label{tbl:question_examples} Example questions from OTTA and Spider. We grouped the examples by the hardness scores. The examples are for the Chinook domain, which is an online music store database.}
\end{minipage}}

\end{table*}

\noindent {\bf Examples.} In Table \ref{tbl:question_examples}, we show examples of questions from OTTA compared to questions from Spider. The examples show that the quality of the questions is similar. The \emph{easy} questions in both datasets are often only simple filtering questions on one table. \emph{Medium} complexity questions include join operations and filters. \emph{Hard} questions in both datasets include join operations and aggregation operations such as finding the maximum or computing the average. The biggest difference is in the \emph{Extra} complexity. There Spider focuses more on subqueries in the where clause. OTTA, on the other hand, focuses more on larger join paths, which are typical for real-world database queries as well as group-by operations and aggregations.

\section{Baseline Systems}
\noindent {\bf Baseline model.} As baseline model for OTs from NL questions, we follow the \emph{Syntactic Neural Model for Code Generation} by \cite{yin-neubig-2017-syntactic}, which we refer to as \emph{Grammar-RNN}\footnote{The IR-Net \cite{guo-etal-2019-towards} is also based on the Grammar-RNN. During the time of writing this paper, IR-Net was ranked second on the Spider leader board.}. This model is based on an encoder-decoder architecture that learns to generate a sequence of production rules of an arbitrary grammar, which in turn produces the query for a given question. For a more detailed discussion on this architecture, we refer the reader to \cite{yin-neubig-2017-syntactic}. In our case, it learns to generate the rules defined in Figure \ref{fig:grammar} for a given question in natural language. Based on the generated list of rules, an OT is created.

We train the model in two phases - a \emph{pre-training phase} and a \emph{supervised phase}. In the pre-training phase, we train a grammar-autoencoder on large amounts of randomly sampled OTs. In the supervised phase, we replace the grammar-encoder by a text encoder and train on the labelled dataset, i.e., the samples with NL question and corresponding OT. 


\noindent {\bf Encoder.} For the NL question, we use a standard Gated-Recurrent Unit (GRU) \cite{chung2014empirical} to encode the question. If $w_i$ denotes the representation of the i-th token in the question, then the encoder produces a corresponding hidden state $h^{E}_i$. Let $H^E \in \mathbb{R}^{N \times h}$ denote the concatenation of all hidden states produced by the GRU for one question, where $N$ is the number of tokens and $h$ the size of the hidden state.  

\noindent {\bf Decoder.} The decoder learns to generate a sequence of production rules with which a tree $y$ is generated for a given encoding $x$ of the NL question. The generation process is formalized as: 
\begin{equation}
    p(y \mid x) = \prod_{t = 1}^T p(a_t\mid x, a_{<t}, a_{p_t})
\end{equation}

$a_t$ is the action taken at time t, $a_{<t}$ are the actions taken before time t, $ a_{p_t}$ are the parent actions taken, and $x$ is the encoded input question. 
There are two different types of rules that the model applies during decoding: 1) If the current rule generates a non-terminal symbol, then \emph{ApplyRule[r]} is executed, which applies a production rule to the current tree. 2) If the next symbol is a terminal, then \emph{GenToken[v]} is applied, which selects the token from a vocabulary. In our case, we have different types of tokens to be generated: table-names, attribute-names and filter operations. 
Similar to \emph{Grammar-RNN}, we implement the decoder using a recurrent neural network, where the internal state is given by: 
\begin{equation}
    h_t = \textit{GRU}([a_{t-1}: c_t : a_{p_t} : n_{f_t}], \widetilde{h}_{t-1})
\end{equation}
$n_{f_t}$ is the embedding of the current node type (e.g. average, union, ...), $c_t$ is a context vector that is computed by applying soft-attention over the input hidden states $H^E$, and $h_{t-1}$ is the hidden vector of the last state. In contrast to \cite{yin-neubig-2017-syntactic}, we apply attention based on \cite{luong-etal-2015-effective}, where $\widetilde{h}_{t-1} = tanh(W_c[h_{t-1} : c_t])$.

For the selection of the terms, we have four output matrices $W_R, W_T, W_A, W_C$, where $W_R$ encodes the grammar rules (i.e. for the non-terminal symbols), and $W_T, W_A, W_C$ encode the table names, attributes and comparison operations, respectively. Depending on the current frontier node, the next output is computed by:
\begin{equation}
   a_t = \textit{argmax}(\textit{softmax}(W_R * h_t))
\end{equation}
\noindent {\bf Grammar Encoder. } The tree encoder, which we use for the pre-training, is based on the same GRU architecture as the decoder. The hidden states for each rule are computed by: 
\begin{equation}
    h_t = \textit{GRU}([a_{t-1}: a_{p_t} : n_{f_t}], h_{t-1})
\end{equation}
In contrast to the encoder, there is no context vector $c_t$. Moreover, $h_{t-1}$ is the last hidden state computed by the GRU. The output of the encoder is a sequence of all states: $H^R \in \mathbb{R}^{R \times h}$, where R denotes the number of rules in the encoded tree.

\noindent {\bf Token Attention.} A straight-forward method to include the explicit token alignment, which is created in the second annotation phase, is to force the attention mechanism to learn the alignment. For this, we add an extra loss function, which computes the binary cross entropy for each attention weight. 

More formally, let $\alpha_t = \textit{softmax}(h_{t - 1} H^E) \in \mathbb{R}^{N}$ be the attention weights computed for timestep t (during the pre-training phase $H^E$ is replaced by $H^R$). Then let $\alpha_t^{(i)}$ be the attention weight for the i-th token. For each token we add the loss 
\begin{equation}
g_i*log(\alpha_t^{(i)}) + (1 - g_i)*log(1 - \alpha_t^{(i)}),
\end{equation}

where $g_i \in [0,1]$ denotes if the token is assigned to the current node or not.

\section{Results}
We now report the results of our model. The details of the experimental setup can be found in Appendix \ref{sec:appendix_setup}. Each experiment is repeated five times with different random seeds.
Table \ref{table:complexity-score} shows the precision of the \emph{Grammar-RNN} on the 5 datasets of \CORPUSNAME. The \emph{precision} is defined as the exact result set matching between the gold standard query and the generated query. Furthermore, the table shows the average precision for each query complexity category. The column "Weighted Avg." refers to the mean average precision over all queries irrespective of the query complexity category.

\noindent {\bf Precision.} For all the databases, except FORMULA-1, the model achieves a precision between $45.1\%$ and $47.5\%$. For FORMULA-1 the model only achieves a score of $26.3\%$. This could be explained by the fact that the FORMULA-1 database contains 93 different attributes, and our data only covers 42 of these attributes. Furthermore, each attribute appears only 17.1 times per query on average. In contrast, for the COLLEGE database the attributes appear in 56 queries on average. Thus, it is harder for the model to learn attributes, which do not appear often in the training set. 
For most of the databases, the model cannot handle the \emph{extra hard} questions, which often contain multiple joins, aggregations, and/or group by operators. Note that without the pre-training phase, the scores drop by a large margin. For instance, the scores for Moviedata drop below 30\% precision.

\begin{table}[h!]
\vspace{-1mm}
\begin{center}
\begin{small}
\resizebox{.5\textwidth}{!}{
\begin{tabular}{l||cccc|c} 
\hline
\textsc{} & Easy& Medium & Hard & Extra Hard & Weighted Avg.\\
\hline 
\textsc{Moviedata} & 0.645 & 0.619 & 0.437 & 0.108 & 0.475\\
\textsc{Chinook} & 0.610 & 0.442 & 0.396 &  0.482 & 0.473\\
\textsc{College} & 0.525  & 0.739 & 0.294 & 0.077 & 0.468\\
\textsc{Driving} School & 0.518 & 0.272 & 0.611 &  0.187 & 0.451\\
\textsc{Formula} 1 & 0.355  & 0.075 & 0.0 & 0.0  & 0.263\\
\hline
\end{tabular}
}
\end{small}
\\~\\
  \vspace{1mm}
\end{center}
\caption{Precision of queries against our 5 datasets according to query complexity. "Weighted Avg." refers to the mean average precision over all queries irrespective of the query complexity category.}
\label{table:complexity-score}
\end{table}

\begin{figure}[h!]
	\begin{center}
		\includegraphics[width=0.45\textwidth]{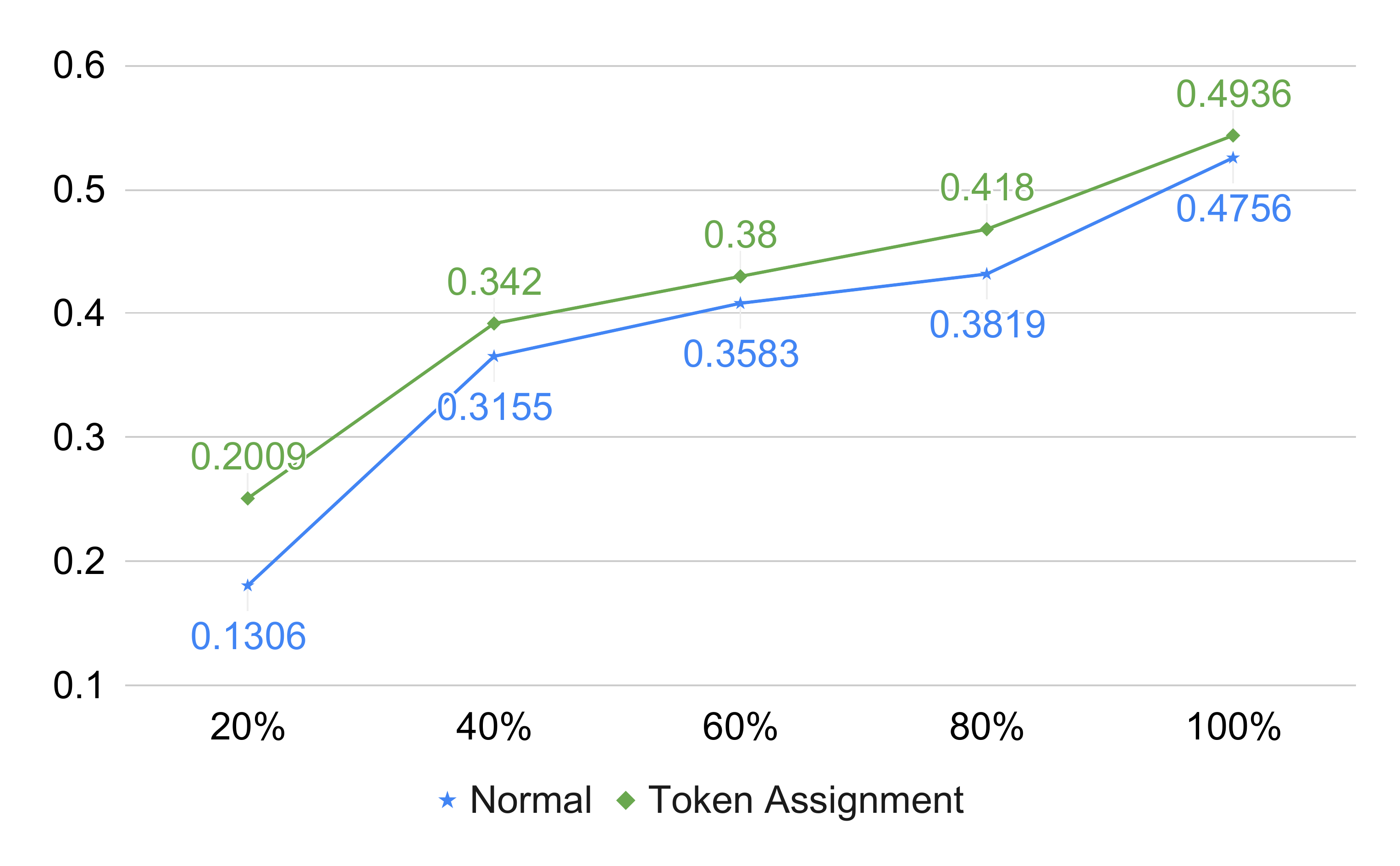}
	\end{center}
	\caption{Learning curve for 20\%, 40\%, 60\%, 80\%, and 100\% of the data on database \emph{MOVIEDATA} as part of \CORPUSNAME. We compare the scores for the training with and without token alignment.}
\label{fig:learning-curve}
\end{figure}

\noindent {\bf Benefit from Token Assignments.}  We now evaluate whether the token assignments can help to train better models. Figure \ref{fig:learning-curve} displays the learning curves for the MOVIEDATA database \emph{ with and without} the token assignment. The model is trained with 20\%, 40\%, 60\%, 80\%, and 100\% of the data. The results show that using the token assignment increases the scores by around $2\%$. In the case of  20\% training data, the gain is even as high as $7\%$, thus showing that the model can benefit from the additional information that is provided in the token assignments. 

\section{Conclusion}
In this paper, we introduced a fast annotation procedure to create NL queries and corresponding database queries (in our case, Operation Trees). Our procedure more than triples the velocity of annotation in comparison to previous methods, while ensuring a larger variety of different types of queries and covering a larger part of the underlying databases. Furthermore, our procedure allows a fine-grained alignment of tokens to operations.
We then used our new method to generate \emph{\CORPUSNAME}, a novel corpus for semantic parsing based on operation trees in combination with token assignments. Generating this corpus was more time- and cost-efficient than with previous approaches. Our statistical analysis showed that the corpus yields a higher coverage of attributes in the databases and more complex natural language questions than other existing methods.  
Furthermore, we implemented a baseline system for automatically generating OTs from NL queries. This baseline achieves scores of up to $48\%$ precision, which are already reasonable while also leaving large potential for improvement in future research. Finally, we showed that the inclusion of the token alignment results in an increase of precision of up to $7\%$. 

Based on these results, we will explore ways to leverage the token assignment to domain adaption and few-shot learning. We also plan to enhance the annotation process by automatically generating proposals for the NL questions and token assignments and letting the annotators only perform corrections. We hope that this increases annotation efficiency even more.

\section{Acknowledgements}
This work has been partially funded by the LIHLITH project supported by the EU ERA-Net CHIST-ERA; the Swiss National Science Foundation [20CH21\_174237]; the Agencia Estatal de Investigación (AEI, Spain) project PCIN-2017-118; the INODE project supported by the European Union’s Horizon 2020 research and innovation program under grant agreement No 863410.

\bibliography{acl2020}

\begin{thebibliography}{33}
\expandafter\ifx\csname natexlab\endcsname\relax\def\natexlab#1{#1}\fi

\bibitem[{Affolter et~al.(2019)Affolter, Stockinger, and
  Bernstein}]{Affolter2019}
Katrin Affolter, Kurt Stockinger, and Abraham Bernstein. 2019.
\newblock \href {https://doi.org/10.1007/s00778-019-00567-8} {A comparative
  survey of recent natural language interfaces for databases}.
\newblock \emph{The VLDB Journal}, 28(5):793--819.

\bibitem[{Basik et~al.(2018)Basik, H{\"a}ttasch, Ilkhechi, Usta, Ramaswamy,
  Utama, Weir, Binnig, and Cetintemel}]{basik2018dbpal}
Fuat Basik, Benjamin H{\"a}ttasch, Amir Ilkhechi, Arif Usta, Shekar Ramaswamy,
  Prasetya Utama, Nathaniel Weir, Carsten Binnig, and Ugur Cetintemel. 2018.
\newblock Dbpal: A learned nl-interface for databases.
\newblock In \emph{Proceedings of the 2018 International Conference on
  Management of Data}, pages 1765--1768. ACM.

\bibitem[{Bast and Haussmann(2015)}]{bast2015more}
Hannah Bast and Elmar Haussmann. 2015.
\newblock More accurate question answering on freebase.
\newblock In \emph{Proceedings of the 24th ACM International on Conference on
  Information and Knowledge Management}, pages 1431--1440. ACM.

\bibitem[{Blunschi et~al.(2012)Blunschi, Jossen, Kossmann, Mori, and
  Stockinger}]{blunschi2012soda}
Lukas Blunschi, Claudio Jossen, Donald Kossmann, Magdalini Mori, and Kurt
  Stockinger. 2012.
\newblock Soda: Generating sql for business users.
\newblock \emph{Proceedings of the VLDB Endowment}, 5(10):932--943.

\bibitem[{Bojanowski et~al.(2017)Bojanowski, Grave, Joulin, and
  Mikolov}]{bojanowski-etal-2017-enriching}
Piotr Bojanowski, Edouard Grave, Armand Joulin, and Tomas Mikolov. 2017.
\newblock \href {https://doi.org/10.1162/tacl_a_00051} {Enriching word vectors
  with subword information}.
\newblock \emph{Transactions of the Association for Computational Linguistics},
  5:135--146.

\bibitem[{Cheng et~al.(2019)Cheng, Reddy, Saraswat, and
  Lapata}]{cheng2019learning}
Jianpeng Cheng, Siva Reddy, Vijay Saraswat, and Mirella Lapata. 2019.
\newblock Learning an executable neural semantic parser.
\newblock \emph{Computational Linguistics}, 45(1):59--94.

\bibitem[{Chung et~al.(2014)Chung, Gulcehre, Cho, and
  Bengio}]{chung2014empirical}
Junyoung Chung, Caglar Gulcehre, KyungHyun Cho, and Yoshua Bengio. 2014.
\newblock Empirical evaluation of gated recurrent neural networks on sequence
  modeling.
\newblock \emph{arXiv preprint arXiv:1412.3555}.

\bibitem[{Covington and McFall(2010)}]{doi:10.1080/09296171003643098}
Michael~A. Covington and Joe~D. McFall. 2010.
\newblock \href {https://doi.org/10.1080/09296171003643098} {Cutting the
  gordian knot: The moving-average type–token ratio (mattr)}.
\newblock \emph{Journal of Quantitative Linguistics}, 17(2):94--100.

\bibitem[{Dahl et~al.(1994)Dahl, Bates, Brown, Fisher, Hunicke-Smith, Pallett,
  Pao, Rudnicky, and Shriberg}]{Dahl:1994:ESA:1075812.1075823}
Deborah~A. Dahl, Madeleine Bates, Michael Brown, William Fisher, Kate
  Hunicke-Smith, David Pallett, Christine Pao, Alexander Rudnicky, and
  Elizabeth Shriberg. 1994.
\newblock \href {https://doi.org/10.3115/1075812.1075823} {Expanding the scope
  of the atis task: The atis-3 corpus}.
\newblock In \emph{Proceedings of the Workshop on Human Language Technology},
  HLT '94, pages 43--48, Stroudsburg, PA, USA. Association for Computational
  Linguistics.

\bibitem[{Damljanovic et~al.(2010)Damljanovic, Agatonovic, and
  Cunningham}]{damljanovic2010natural}
Danica Damljanovic, Milan Agatonovic, and Hamish Cunningham. 2010.
\newblock Natural language interfaces to ontologies: Combining syntactic
  analysis and ontology-based lookup through the user interaction.
\newblock In \emph{Extended Semantic Web Conference}, pages 106--120. Springer.

\bibitem[{Dubey et~al.(2019)Dubey, Banerjee, Abdelkawi, and
  Lehmann}]{dubey2019lc}
Mohnish Dubey, Debayan Banerjee, Abdelrahman Abdelkawi, and Jens Lehmann. 2019.
\newblock Lc-quad 2.0: A large dataset for complex question answering over
  wikidata and dbpedia.
\newblock In \emph{International Semantic Web Conference}, pages 69--78.
  Springer.

\bibitem[{Ferr{\'e}(2017)}]{ferre2017sparklis}
S{\'e}bastien Ferr{\'e}. 2017.
\newblock Sparklis: an expressive query builder for sparql endpoints with
  guidance in natural language.
\newblock \emph{Semantic Web}, 8(3):405--418.

\bibitem[{Finegan-Dollak et~al.(2018)Finegan-Dollak, Kummerfeld, Zhang,
  Ramanathan, Sadasivam, Zhang, and Radev}]{acl18sql}
Catherine Finegan-Dollak, Jonathan~K. Kummerfeld, Li~Zhang, Karthik Ramanathan,
  Sesh Sadasivam, Rui Zhang, and Dragomir Radev. 2018.
\newblock \href {http://www.aclweb.org/anthology/P18-1033.pdf} {Improving
  text-to-sql evaluation methodology}.
\newblock In \emph{Proceedings of the 56th Annual Meeting of the Association
  for Computational Linguistics (Volume 1: Long Papers)}, pages 351--360,
  Melbourne, Victoria, Australia.

\bibitem[{Giordani and Moschitti(2012)}]{Giordani:2012:AGR:3069470.3069477}
Alessandra Giordani and Alessandro Moschitti. 2012.
\newblock \href {https://doi.org/10.1007/978-3-642-45260-4_5} {Automatic
  generation and reranking of sql-derived answers to nl questions}.
\newblock In \emph{Proceedings of the Second International Conference on
  Trustworthy Eternal Systems via Evolving Software, Data and Knowledge},
  EternalS'12, pages 59--76, Berlin, Heidelberg. Springer-Verlag.

\bibitem[{Guo et~al.(2019)Guo, Zhan, Gao, Xiao, Lou, Liu, and
  Zhang}]{guo-etal-2019-towards}
Jiaqi Guo, Zecheng Zhan, Yan Gao, Yan Xiao, Jian-Guang Lou, Ting Liu, and
  Dongmei Zhang. 2019.
\newblock \href {https://doi.org/10.18653/v1/P19-1444} {Towards complex
  text-to-{SQL} in cross-domain database with intermediate representation}.
\newblock In \emph{Proceedings of the 57th Annual Meeting of the Association
  for Computational Linguistics}, pages 4524--4535, Florence, Italy.
  Association for Computational Linguistics.

\bibitem[{Iyer et~al.(2017{\natexlab{a}})Iyer, Konstas, Cheung, Krishnamurthy,
  and Zettlemoyer}]{iyer2017learning}
Srinivasan Iyer, Ioannis Konstas, Alvin Cheung, Jayant Krishnamurthy, and Luke
  Zettlemoyer. 2017{\natexlab{a}}.
\newblock Learning a neural semantic parser from user feedback.
\newblock In \emph{Proceedings of the 55th Annual Meeting of the Association
  for Computational Linguistics (Volume 1: Long Papers)}, pages 963--973.

\bibitem[{Iyer et~al.(2017{\natexlab{b}})Iyer, Konstas, Cheung, Krishnamurthy,
  and Zettlemoyer}]{iyer-etal-2017-learning}
Srinivasan Iyer, Ioannis Konstas, Alvin Cheung, Jayant Krishnamurthy, and Luke
  Zettlemoyer. 2017{\natexlab{b}}.
\newblock \href {https://doi.org/10.18653/v1/P17-1089} {Learning a neural
  semantic parser from user feedback}.
\newblock In \emph{Proceedings of the 55th Annual Meeting of the Association
  for Computational Linguistics (Volume 1: Long Papers)}, pages 963--973,
  Vancouver, Canada. Association for Computational Linguistics.

\bibitem[{Kingma and Ba(2014)}]{kingma2014adam}
Diederik~P Kingma and Jimmy Ba. 2014.
\newblock Adam: A method for stochastic optimization.
\newblock \emph{arXiv preprint arXiv:1412.6980}.

\bibitem[{Li and Jagadish(2014)}]{li2014constructing}
Fei Li and HV~Jagadish. 2014.
\newblock Constructing an interactive natural language interface for relational
  databases.
\newblock \emph{Proceedings of the VLDB Endowment}, 8(1):73--84.

\bibitem[{Liu et~al.(2019)Liu, Fang, Liu, Chen, Jian-Guang, and
  Li}]{liu2019leveraging}
Haoyan Liu, Lei Fang, Qian Liu, Bei Chen, LOU Jian-Guang, and Zhoujun Li. 2019.
\newblock Leveraging adjective-noun phrasing knowledge for comparison relation
  prediction in text-to-sql.
\newblock In \emph{Proceedings of the 2019 Conference on Empirical Methods in
  Natural Language Processing and the 9th International Joint Conference on
  Natural Language Processing (EMNLP-IJCNLP)}, pages 3506--3511.

\bibitem[{Luong et~al.(2015)Luong, Pham, and
  Manning}]{luong-etal-2015-effective}
Thang Luong, Hieu Pham, and Christopher~D. Manning. 2015.
\newblock \href {https://doi.org/10.18653/v1/D15-1166} {Effective approaches to
  attention-based neural machine translation}.
\newblock In \emph{Proceedings of the 2015 Conference on Empirical Methods in
  Natural Language Processing}, pages 1412--1421, Lisbon, Portugal. Association
  for Computational Linguistics.

\bibitem[{Popescu et~al.(2003)Popescu, Etzioni, and
  Kautz}]{Popescu:2003:TTN:604045.604070}
Ana-Maria Popescu, Oren Etzioni, and Henry Kautz. 2003.
\newblock \href {https://doi.org/10.1145/604045.604070} {Towards a theory of
  natural language interfaces to databases}.
\newblock In \emph{Proceedings of the 8th International Conference on
  Intelligent User Interfaces}, IUI '03, pages 149--157, New York, NY, USA.
  ACM.

\bibitem[{Saha et~al.(2016)Saha, Floratou, Sankaranarayanan, Minhas, Mittal,
  and {\"O}zcan}]{saha2016athena}
Diptikalyan Saha, Avrilia Floratou, Karthik Sankaranarayanan, Umar~Farooq
  Minhas, Ashish~R Mittal, and Fatma {\"O}zcan. 2016.
\newblock Athena: an ontology-driven system for natural language querying over
  relational data stores.
\newblock \emph{Proceedings of the VLDB Endowment}, 9(12):1209--1220.

\bibitem[{Simitsis et~al.(2008)Simitsis, Koutrika, and
  Ioannidis}]{simitsis2008precis}
Alkis Simitsis, Georgia Koutrika, and Yannis Ioannidis. 2008.
\newblock Pr{\'e}cis: from unstructured keywords as queries to structured
  databases as answers.
\newblock \emph{The VLDB Journal—The International Journal on Very Large Data
  Bases}, 17(1):117--149.

\bibitem[{Song et~al.(2015)Song, Schilder, Smiley, Brew, Zielund, Bretz,
  Martin, Dale, Duprey, Miller et~al.}]{song2015tr}
Dezhao Song, Frank Schilder, Charese Smiley, Chris Brew, Tom Zielund, Hiroko
  Bretz, Robert Martin, Chris Dale, John Duprey, Tim Miller, et~al. 2015.
\newblock Tr discover: A natural language interface for querying and analyzing
  interlinked datasets.
\newblock In \emph{International Semantic Web Conference}, pages 21--37.
  Springer.

\bibitem[{Tang and Mooney(2000)}]{tang-mooney-2000-automated}
Lappoon~R. Tang and Raymond~J. Mooney. 2000.
\newblock \href {https://doi.org/10.3115/1117794.1117811} {Automated
  construction of database interfaces: Intergrating statistical and relational
  learning for semantic parsing}.
\newblock In \emph{2000 Joint {SIGDAT} Conference on Empirical Methods in
  Natural Language Processing and Very Large Corpora}, pages 133--141, Hong
  Kong, China. Association for Computational Linguistics.

\bibitem[{Wang et~al.(2015)Wang, Berant, and Liang}]{wang-etal-2015-building}
Yushi Wang, Jonathan Berant, and Percy Liang. 2015.
\newblock \href {https://doi.org/10.3115/v1/P15-1129} {Building a semantic
  parser overnight}.
\newblock In \emph{Proceedings of the 53rd Annual Meeting of the Association
  for Computational Linguistics and the 7th International Joint Conference on
  Natural Language Processing (Volume 1: Long Papers)}, pages 1332--1342,
  Beijing, China. Association for Computational Linguistics.

\bibitem[{Yaghmazadeh et~al.(2017)Yaghmazadeh, Wang, Dillig, and
  Dillig}]{Yaghmazadeh:2017:SQS:3152284.3133887}
Navid Yaghmazadeh, Yuepeng Wang, Isil Dillig, and Thomas Dillig. 2017.
\newblock \href {https://doi.org/10.1145/3133887} {Sqlizer: Query synthesis
  from natural language}.
\newblock \emph{Proc. ACM Program. Lang.}, 1(OOPSLA):63:1--63:26.

\bibitem[{Yin and Neubig(2017)}]{yin-neubig-2017-syntactic}
Pengcheng Yin and Graham Neubig. 2017.
\newblock \href {https://doi.org/10.18653/v1/P17-1041} {A syntactic neural
  model for general-purpose code generation}.
\newblock In \emph{Proceedings of the 55th Annual Meeting of the Association
  for Computational Linguistics (Volume 1: Long Papers)}, pages 440--450,
  Vancouver, Canada. Association for Computational Linguistics.

\bibitem[{Yu et~al.(2018)Yu, Zhang, Yang, Yasunaga, Wang, Li, Ma, Li, Yao,
  Roman, Zhang, and Radev}]{yu-etal-2018-spider}
Tao Yu, Rui Zhang, Kai Yang, Michihiro Yasunaga, Dongxu Wang, Zifan Li, James
  Ma, Irene Li, Qingning Yao, Shanelle Roman, Zilin Zhang, and Dragomir Radev.
  2018.
\newblock \href {https://doi.org/10.18653/v1/D18-1425} {{S}pider: A large-scale
  human-labeled dataset for complex and cross-domain semantic parsing and
  text-to-{SQL} task}.
\newblock In \emph{Proceedings of the 2018 Conference on Empirical Methods in
  Natural Language Processing}, pages 3911--3921, Brussels, Belgium.
  Association for Computational Linguistics.

\bibitem[{Zelle and Mooney(1996)}]{Zelle:1996:LPD:1864519.1864543}
John~M. Zelle and Raymond~J. Mooney. 1996.
\newblock \href {http://dl.acm.org/citation.cfm?id=1864519.1864543} {Learning
  to parse database queries using inductive logic programming}.
\newblock In \emph{Proceedings of the Thirteenth National Conference on
  Artificial Intelligence - Volume 2}, AAAI'96, pages 1050--1055. AAAI Press.

\bibitem[{Zheng et~al.(2017)Zheng, Cheng, Zou, Yu, and Zhao}]{zheng2017natural}
Weiguo Zheng, Hong Cheng, Lei Zou, Jeffrey~Xu Yu, and Kangfei Zhao. 2017.
\newblock Natural language question/answering: Let users talk with the
  knowledge graph.
\newblock In \emph{Proceedings of the 2017 ACM on Conference on Information and
  Knowledge Management}, pages 217--226. ACM.

\bibitem[{Zhong et~al.(2017)Zhong, Xiong, and Socher}]{zhongSeq2SQL2017}
Victor Zhong, Caiming Xiong, and Richard Socher. 2017.
\newblock Seq2sql: Generating structured queries from natural language using
  reinforcement learning.
\newblock \emph{CoRR}, abs/1709.00103.

\end{thebibliography}
\bibliographystyle{acl_natbib}
\clearpage

\appendix

\section{Experimental Setup}
\label{sec:appendix_setup}
\paragraph{Preprocessing} We use Spacy\footnote{https://spacy.io/} to tokenize the NL questions in \CORPUSNAME. In order to find the entities for the constraints, we employ simple string matching. With this, we find $96\%$ of all entities. Thus, the generated OTs are executable, and we can compare the results of the generated OT to the results by the gold-standard OT from the corpus.

\paragraph{Model Configuration.}  For our model, we chose a hidden layer of $h=256$ dimensions. We optimize using the \emph{Adam } \cite{kingma2014adam} optimizer with the standard values. We let the model train using early stopping with a patience on the validation loss, where the validation set is the left-out fold in 5 fold cross-validation. For the word embeddings, we use the pre-trained \emph{FastText} embeddings \cite{bojanowski-etal-2017-enriching}, which are refined during the training phase.

\section{Query Complexity}
The tables below show more details of the coverage (see Tables \ref{table:table-coverage} and \ref{table:attributes-coverage}) and the average number of joins per query (see Table \ref{table:amount-joins}).

\begin{table*}[h!]
\vspace{-1mm}
\begin{center}
\begin{small}
\resizebox{.8\textwidth}{!}{
\begin{tabular}{l||ccccc} 
\hline
& \textsc{Total} & \textsc{CHINOOK} & \textsc{COLLEGE} & \textsc{DRIVING SCHOOL} & \textsc{FORMULA\_1}\\
\hline 
\textsc{Spider}  &    0. 917 (0.87)     &   0.727    &    0.909   &    1           &    0.692    \\
\textsc{Our Dataset} &  0.949       & 1   &   0.818    &        1      &   0.923    \\
\hline
\end{tabular}
}
\end{small}
  \vspace{1mm}
\end{center}
\caption{Table Coverage, in \% to total amount of existing tables. Our dataset shows better table coverage, except for one database (college), where the coverage differs by one table. The biggest improvement in coverage was achieved on the database formula\_1, which is also the most complex database with the largest amount of tables. The number in braces indicates the average table coverage for the databases with more than 5 tables.}
\label{table:table-coverage}
\end{table*}

\begin{table*}[h!]
\vspace{-1mm}
\begin{center}
\begin{small}
\resizebox{.8\textwidth}{!}{
\begin{tabular}{l||ccccc} 
\hline
& \textsc{Total} & \textsc{CHINOOK} & \textsc{COLLEGE} & \textsc{DRIVING SCHOOL} & \textsc{FORMULA\_1}\\
\hline 
\textsc{Spider}  &     0.621 (0.496)    &    0.354   &   0.383    &  0.730   &   0.221    \\
\textsc{Our Dataset} &    0.544    &  0.584  &   0.384   &     0.756     &    0.442    \\
\hline
\end{tabular}
}
\end{small}
\\~\\
  \vspace{1mm}
\end{center}
\caption{Attribute Coverage. Our method gives better attribute coverage in particular for larger datasets, for instance, FORMULA\_1.  The number in braces indicates the average attribute coverage for the databases with more than 5 tables.}
\label{table:attributes-coverage}
\end{table*}

\begin{table*}[h!]
\vspace{-1mm}
\begin{center}
\begin{small}
\resizebox{.8\textwidth}{!}{
\begin{tabular}{l||ccccc} 
\hline
& \textsc{Total} & \textsc{CHINOOK} & \textsc{COLLEGE} & \textsc{DRIVING SCHOOL} & \textsc{FORMULA\_1}\\
\hline 
\textsc{Spider}  &   0.504     &    0.667   &      0.412 &  0.441   &   0.925   \\
\textsc{Our Dataset} &   1.15     &  0.95  &   1.18   &   0.837    &   1.2   \\
\hline
\end{tabular}
}
\end{small}
\\~\\
  \vspace{1mm}
\end{center}
\caption{Average number of joins per query.}
\label{table:amount-joins}
\end{table*}

\section{Example of Tree Sampling}
\begin{figure*}[h]
	\begin{center}
        \begin{tabular}{@{}c@{}}
		\includegraphics[width=0.9\textwidth]{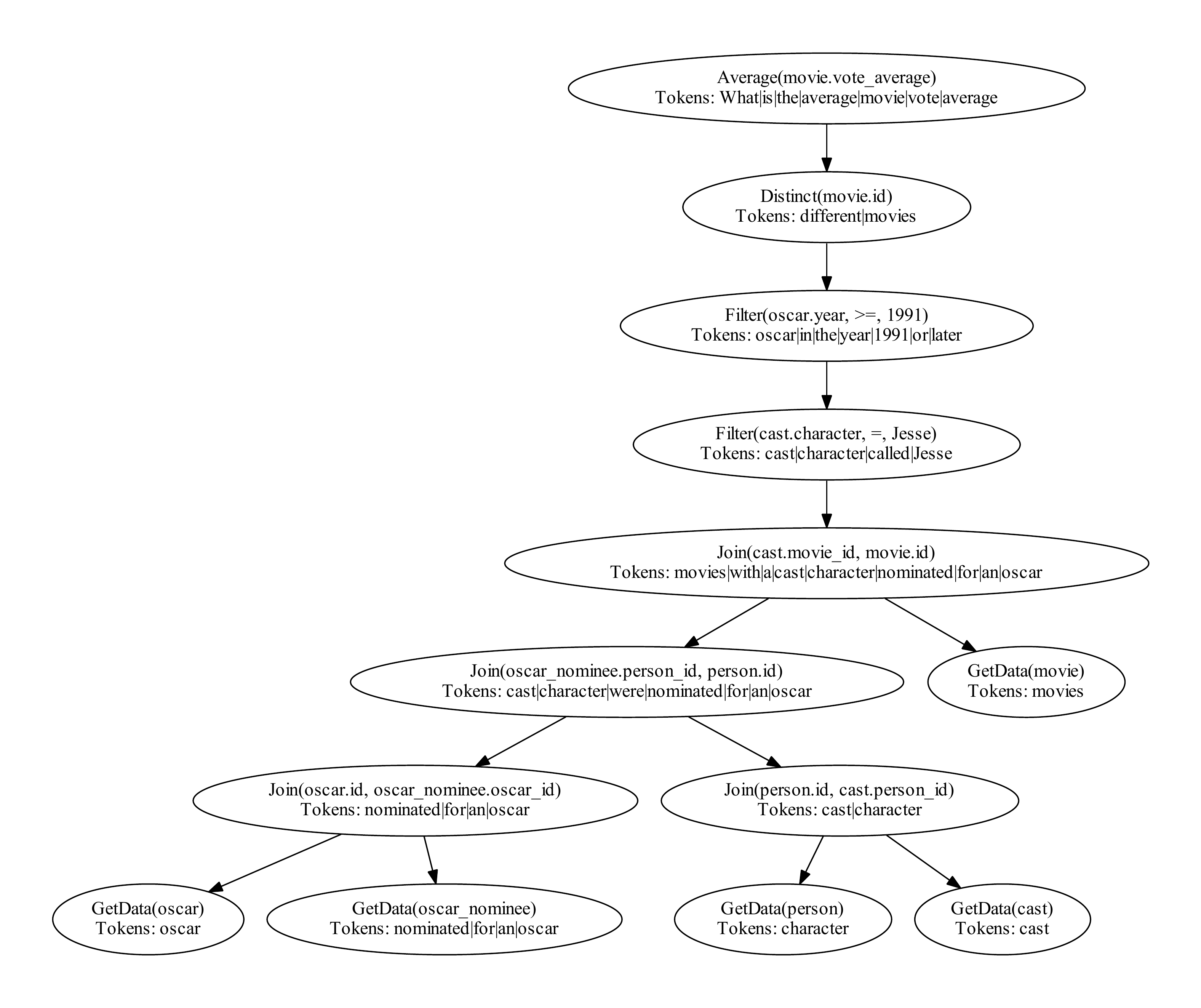}
       \end{tabular}
	\end{center}\vspace{-3mm}
	\captionof{figure}{Example of a randomly sampled tree. The nodes denote the node type with their arguments. The \emph{Tokens} are assigned during the second phase of the annotation process. This tree is the answer to the question: \textit{What is the average movie vote of different movies having an Oscar nominee with a cast character called Jesse and were nominated for an Oscar in the year 1991 or later?}}
\label{fig:tree}
\end{figure*}

Figure \ref{fig:tree} shows a randomly sampled tree. During Phase 1 of the annotation procedure, an annotator associated the tree with the question: \textit{What is the average movie vote of different movies having an Oscar nominee with a cast character called Jesse and were nominated for an Oscar in the year 1991 or later?}. In the second phase of the annotation, the tokens of the questions were associated with the nodes of the tree. The tree is depicted from root to leaves, where the root node is the last operation, and the leave nodes are the \emph{GetData}-nodes. 
Here we describe the tree sampling procedure in more detail with the tree in Figure \ref{fig:tree} as an example.
\begin{itemize}
    \item[1] The query type is selected. There are five different types: list, sum, count, average, and Boolean. In our example, the average was selected. This can be forced manually or randomly sampled.
    \item[2] The result type is selected, which, in this case, is \emph{movie.vote\_average}. This can also be set manually or be sampled at random. Based on the query type, only certain types of results are allowed. More precisely, for average and sum operations, only numeric result types are allowed.
    \item[3] The join path is selected. In the first step, a path length is selected, which can be predefined or randomly sampled. In this case, the path length is set to five. Then, in a second step, a random path of the predefined length is selected. In the current example, the query path is: \emph{movie, cast, person, oscar\_nominee, oscar}. The path always starts with the result type of table.
    \item[4] The set operation is selected among {\it union}, {\it intersection}, or {\it difference}. In this example, there is no set operation. After the operation was selected, a subpath is chosen, on which the set operation is performed. For instance, if we wanted to know the movies where Brad Pitt and George Clooney worked together, then the subpath {\it movie, cast, person} is selected. Finally, two different filters are inserted, one for each actor. 
    \item[5] The {\it group by} operation is selected. First, the operator is selected among {\it sum, average}, or {\it count}. Then, the {\it group by} attribute and the {\it aggregation} attribute are selected. In our example, there is no group by operation.
    \item[6] The aggregation operation is selected among the {\it min} and {\it max} operation. This is relevant for the questions of the type: Which movie has the highest rating. In this example, we have no aggregation operation.
    \item[7] The filters are selected. For this, we define the number of total filters and the maximal number of filters per table. In this case, we set the number of filters equal to 2, and the maximal number of filters per table to one. Then, the appropriate number of attributes is selected randomly alongside the path. In this case, the tables oscar and cast were selected. Then, an attribute is selected, followed by a comparison operator and a value, which is randomly sampled from the database. In our example, we have: oscar.year $\geq$ 1991 and cast.character $=$ Jesse. 
\end{itemize}

\section{Annotation Tool}
\label{sec:appendix_annoation}
\begin{figure*}[h]
	\begin{center}
        \begin{tabular}{@{}c@{}}
		\includegraphics[width=0.95\textwidth]{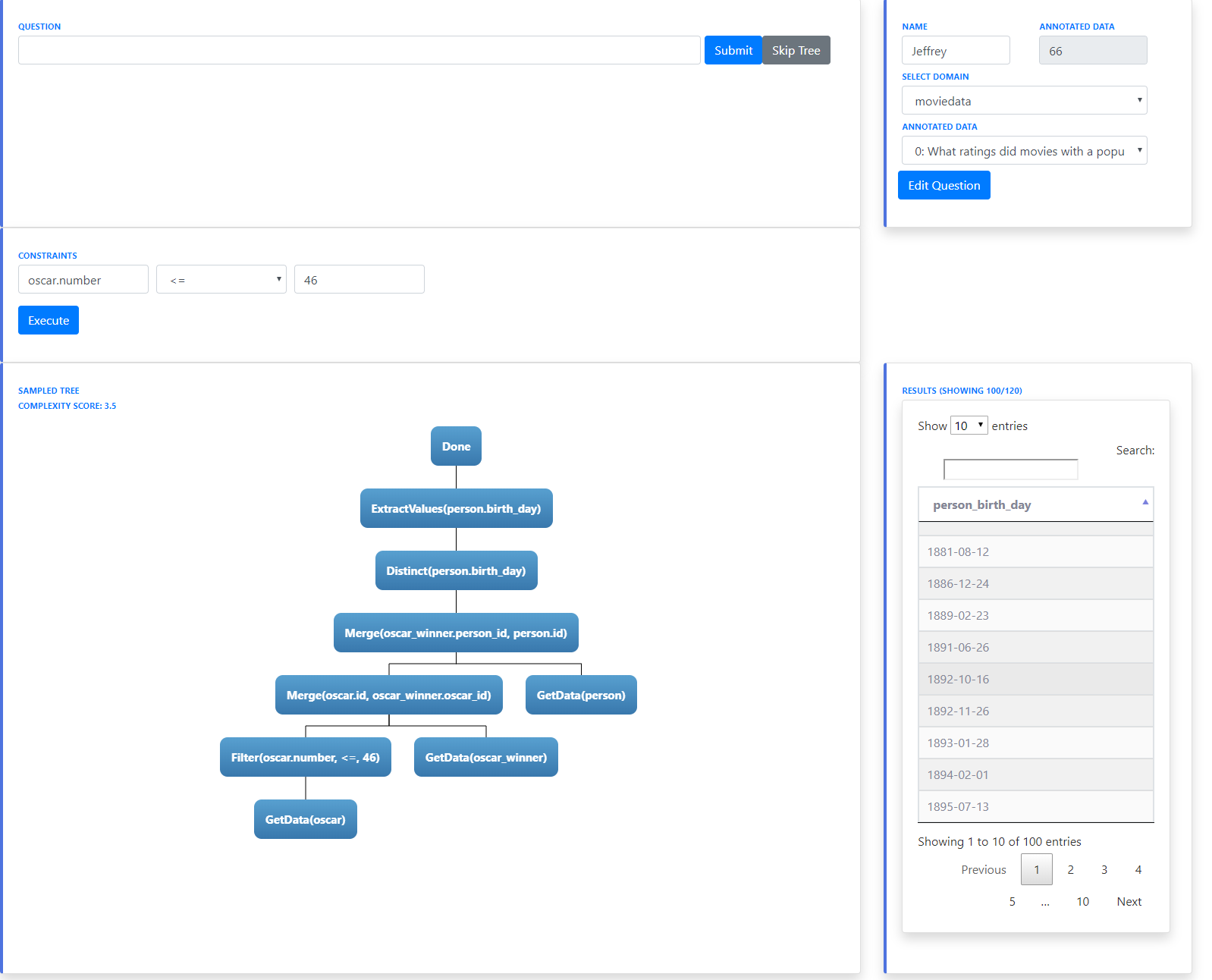} \\
		(a) \vspace{2mm} \\
		\includegraphics[width=0.95\textwidth]{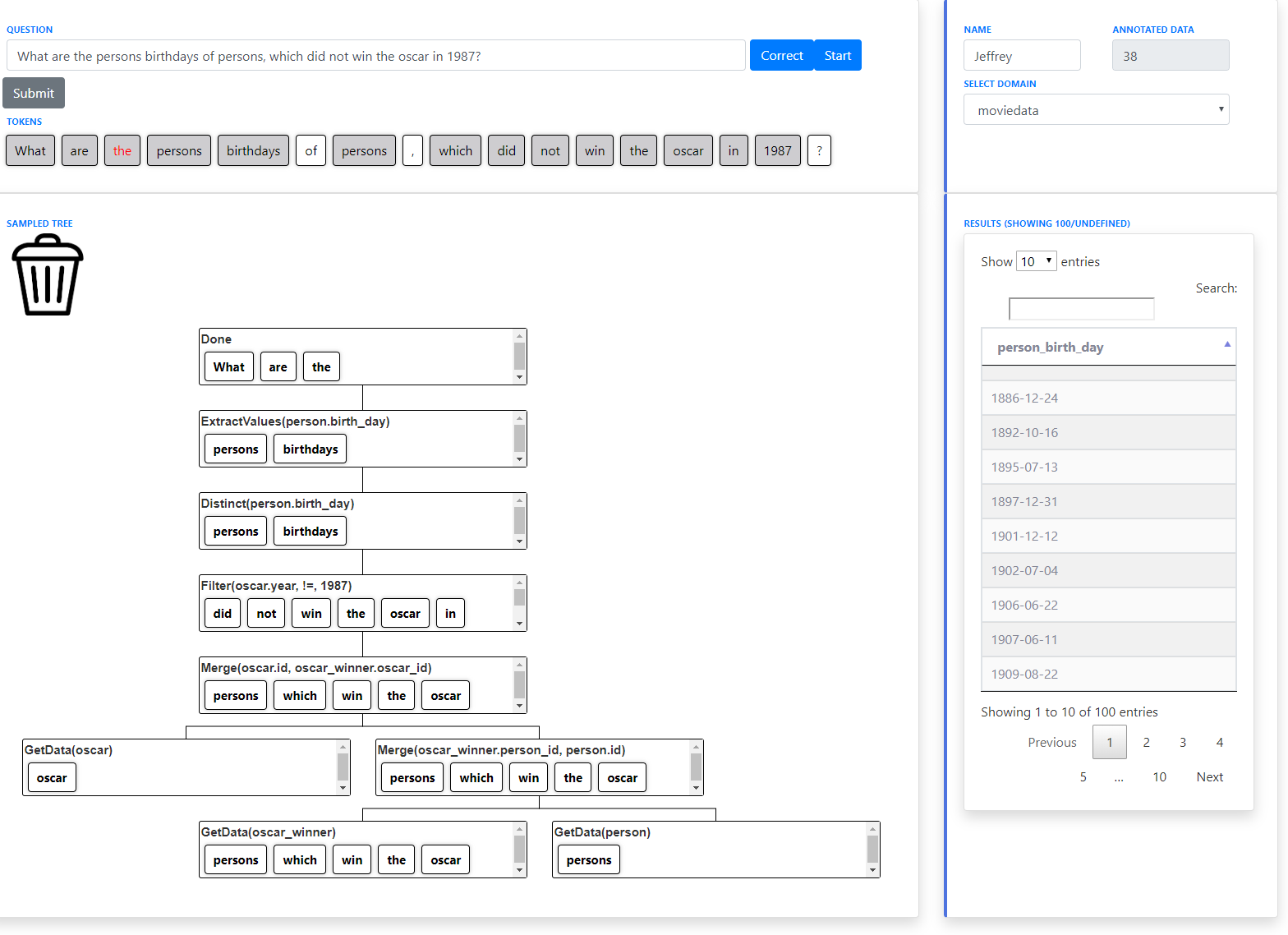} \\
		(b) \\
       \end{tabular}
	\end{center}\vspace{-3mm}
	\captionof{figure}{The annotation tool. (a) The OT and the constraints are shown to the annotators. For each node, the annotators can inspect the result of the execution. The annotators write a question and (b) assign the tokens of the question to the operations.}
\label{fig:tool}
\end{figure*}

The annotation process is performed in two phases: writing an NL question for a given OT, and assigning tokens from the NL question to the nodes within the OT. We have built two user interfaces, one for each phase.  Figure \ref{fig:tool} shows screenshots of both tools.

\paragraph{Phase 1.} In the first phase, the annotators are presented with an OT and the constraints. Their task is to write an appropriate question for the OT. For this, they can adapt the constraints, in case that they are nonsensical. Furthermore, the annotators can access the intermediate results for each node in the tree to better understand what the OT does. In cases where the OT cannot be annotated with an appropriate question, the OT can be skipped. 

\paragraph{Phase 2.} For the second phase (Figure \ref{fig:tool} (b)), the annotators are presented with an OT and an NL question, which was written by another annotator in the previous phase. The task is to correct the question, and then assign the tokens of the NL to the nodes (i.e., operations) in the tree. For this task, the tool guides the annotator from node to node in the OT. Moreover, for each node, the annotator can choose the corresponding tokens. In the final step, the annotators can correct their token assignment using drag-and-drop features.

\paragraph{Guidelines.} In order to have consistent annotations (especially in the second phase), we provided the annotators with extensive tutorial videos. On average, the annotators took 30 minutes to get used to the tool and start to work efficiently. For the first phase, we instructed the annotators to write an appropriate question and gave examples, as well as examples of pitfalls. 

For the second phase, we introduced stricter guidelines, as we noticed that annotators had trouble with this step. Especially, the join operations were not clear to the annotators. Thus, we decided on the following rules:
\begin{itemize}
    \item Table: If the table denotes an entity type (e.g., movie), the tokens that denote this entity type are to be assigned (e.g., "movies"). If the table is a bridge table, which denotes a relationship between entities (e.g., production\_country), then the tokens that denote this relationship are to be assigned to the operation (e.g., "movies", "produced", "in").
    \item Joins: For the join operations, the same guidelines as for the bridge-tables are to be followed. 
    \item Filter: For the filter constraints (e.g., "person\_name= Tom Cruise") the tokens, which represent the constraint, are to be selected (e.g., "by", "Tom", "Cruise").
    \item Query type: For each query type (e.g., count, average, sum, ...), the tokens that correspond or trigger this question type are to be selected (e.g., "How", "many").
\end{itemize}

\paragraph{Annotators.} We recruited 22 annotators, which have a basic understanding of database technologies. We paid each annotator 25\$ per hour. Each annotator was given access to a set of instruction videos as well as a user manual. Furthermore, the annotators could pose questions in a forum. 

\end{document}